\documentclass[journal,twoside,web,twocolumn]{article}

\usepackage{url}
\usepackage{cite}
\usepackage{amsmath,amssymb,amsfonts}
\usepackage{algorithmic}
\usepackage{graphicx}
\usepackage{float}
\usepackage{booktabs}
\usepackage{multirow}
\usepackage{diagbox}
\usepackage{stfloats}
\usepackage{xspace}
\usepackage{textcomp}
\usepackage[T1]{fontenc}

\newcommand{\etal}{\emph{et al.}\xspace}

\newcommand{\mat}[1]{\boldsymbol{#1}} 

\makeatletter

\newcommand{\Rmnum}[1]{\expandafter\@slowromancap\romannumeral #1@}
\makeatother

\def\BibTeX{{\rm B\kern-.05em{\sc i\kern-.025em b}\kern-.08em
		T\kern-.1667em\lower.7ex\hbox{E}\kern-.125emX}}

\title{Multi-task MR Imaging with Iterative Teacher Forcing and Re-weighted Deep Learning}
\author{Kehan Qi
	\and Yu Gong\and
	 Xinfeng Liu\and
	  Xin Liu\and
	  Hairong Zheng\and Shanshan Wang,\textit{Senior Member, IEEE}
	\thanks{This research was partly supported by Scientific and Technical Innovation 2030-"New Generation Artificial Intelligence"  Project (2020AAA0104100, 2020AAA0104105) the National Natural Science Foundation of China (61871371, 81830056), Key-Area Research and Development Program of Guangdong Province (2018B010109009), the Basic Research Program of Shenzhen (JCYJ20180507182400762), Youth Innovation Promotion Association Program of Chinese Academy of Sciences (2019351). (Kehan Qi and Yu Gong are co-first authors. Corresponding author: Shanshan Wang.)}
\thanks{K. Qi, X. Liu, H. Zheng and S. Wang are with Paul C. Lauterbur Research Center for Biomedical Imaging, Shenzhen Institutes of Advanced Technology, Chinese Academy of Sciences, Shenzhen 518055, China; S. Wang is also with Pengcheng Laboratory, Shenzhen, Guangdong (e-mail: kh.qi@siat.ac.cn; ss.wang@siat.ac.cn).}
\thanks{Y. Gong is with the College of Medicine and Biological Information Engineering, Northeastern University, Shenyang 110169, China and also with Paul C. Lauterbur Research Center for Biomedical Imaging, Shenzhen Institutes of Advanced Technology, Chinese Academy of Sciences, Shenzhen 518055, China (e-mail:yu.90n9@gmail.com). X. Liu is with Guizhou Provincial People's hospital, China}}
\date{}

\begin{document}

\maketitle

\begin{abstract}
Noises, artifacts, and loss of information caused by the magnetic resonance (MR) reconstruction may compromise the final performance of the downstream applications. In this paper, we develop a re-weighted multi-task deep learning method to learn prior knowledge from the existing big dataset and then utilize them to assist simultaneous MR reconstruction and segmentation from the under-sampled k-space data. The multi-task deep learning framework is equipped with two network sub-modules, which are integrated and trained by our designed iterative teacher forcing scheme (ITFS) under the dynamic re-weighted loss constraint (DRLC). The ITFS is designed to avoid error accumulation by injecting the fully-sampled data into the training process. The DRLC is proposed to dynamically balance the contributions from the reconstruction and segmentation sub-modules so as to co-prompt the multi-task accuracy. The proposed method has been evaluated on two open datasets and one in vivo in-house dataset and compared to six state-of-the-art methods. Results show that the proposed method possesses encouraging capabilities for simultaneous and accurate MR reconstruction and segmentation.
\end{abstract}

\section{Introduction}
\label{sec:introduction}
Magnetic resonance (MR) imaging is of great value in clinical applications such as medical diagnosis \cite{bruno2019new}, disease staging \cite{taylor2019diagnostic} and clinical research \cite{debette2019clinical} since it can produce highly detailed images with excellent soft tissue contrast. The raw measurements of MR images are Fourier transform coefficients in "k-space" and MR images can be visualized after an inverse Fourier transform of the fully-sampled k-space data. However, the acquisition time of fully-sampled k-space data is long. It is one of the main limitations for the MR extensive applications. Therefore, numerous techniques have been developed for accelerating MR imaging over the last decades, including physics-based fast imaging sequences \cite{oppelt1986fisp}, hardware-based parallel imaging \cite{lustig2010spirit}, signal processing-based image reconstruction from under-sampled k-space data \cite{lustig2008compressed,2014Magnetic} and deep learning-based techniques \cite{wang2016accelerating,2018Deep,2018MoDL,kabkab2018task}. 

In some clinical application scenarios, images are not an end in themselves, but rather a means of access to clinically relevant parameters which are obtained as post-processing steps (e.g., segmentation). However, most of the existing segmentation methods are seen as a post-processing task that is independent of reconstruction. It may be an issue that a lot of information discarded during the reconstruction process may influence the final segmentation performance. The reason is that most of the existing reconstruction methods are designed to take optimal visual quality as the first priority, rather than the specific-task quality.  Therefore, some groups tend to develop task-driven imaging methods for segmentation \cite{adler2018task,corona2019enhancing,ramlau2007mumford,burger2016simultaneous,lauze2017simultaneous,klann2011mumford,klann2011mumfordd}. For instance, Caballero \etal \cite{caballero2014application} proposed an unsupervised brain segmentation method from k-space based on the patch-based sparse modeling and Gaussian mixture modeling. It is a pioneering work for 
task-driven MRI. But its efficiency can still be improved. Recently, many deep learning methods based on convolutional neural networks have been introduced for MR segmentation from k-space data \cite{schlemper2018cardiac, ronneberger2015u,huang2019fr,huang2019brain,sun2019joint}. Schlemper \etal \cite{schlemper2018cardiac} proposed using two neural networks, namely end-to-end synthesis network (SynNet) and latent feature interpolation network (LI-Net) to directly predict segmentation results from under-sampled k-space data. The SynNet uses U-Net \cite{ronneberger2015u} as the backbone to map zero-filling images (inverse Fourier transform of under-sampled k-space) to segmentation maps directly. The LI-Net constructs a latent feature interpolation network with improved results. Huang \etal \cite{huang2019fr} proposed a Joint-FR-Net that consists of a reconstruction module derived from the fast iterative shrinkage-thresholding algorithm (FISTA) and a segmentation module based on U-Net \cite{ronneberger2015u}. Huang \etal \cite{huang2019brain} proposed a unique task-driven attention module that utilizes intermediate segmentation estimation to facilitate image-domain feature extraction from the raw data to bridge the gap between reconstruction and segmentation. Sun \etal \cite{sun2019joint} proposed a unified deep neural network called SegNetMRI with two modules (the reconstruction module and segmentation module). The two modules are pre-trained and fine-tuned with shared reconstruction encoders. The outputs from all segmentation modules are merged into the final segmentation result. All these works \cite{schlemper2018cardiac,huang2019fr,huang2019brain,sun2019joint} have made contributions to prompt the development of task-driven MR imaging. However, existing task-driven methods tend to attribute too much importance to the task (segmentation), which may introduce incorrect reconstruction, such as aliasing or loss of information. Consequently, the aliasing in reconstruction could mislead the network from identifying the correct boundaries. Therefore, it is in need to develop a method for a proper balance and co-promotion of both reconstruction and segmentation accuracy. 

To this end, this paper develops a multi-task MR imaging approach with iterative teacher forcing and re-weighted deep learning. The multi-task deep learning framework is equipped with reconstruction and segmentation  sub-network modules, which are integrated and trained by our designed iterative teacher forcing scheme (ITFS) under the dynamic re-weighted loss constraint (DRLC). Teacher forcing \cite{williams1989learning} scheme is a training technique widely used in the recurrent neural network (RNN) to reduce the training difficulty of the dual module network. We developed the ITFS to replace intermediate reconstruction outputs with the fully-sampled data. It is designed to avoid the bias explosion and over-correction problem caused by the original teacher forcing scheme. To the best of our knowledge, this is the first time teacher forcing concept has been used outside of RNN architecture. The DRLC is proposed to dynamically balance the contributions from the reconstruction and segmentation sub-modules so as to co-prompt the multi-task accuracy. The contributions of this paper are summarized in three aspects as follows:
\begin{enumerate}
	\item[1)] A multi-task MR imaging framework is proposed to derive high-quality image reconstructions and segmented masks simultaneously from undersampled k-space data with iterative teacher forcing and re-weightd deep learning. 
	\item[2)] A dynamic re-weighted loss constraint is designed to help the proposed network to co-prompt the reconstruction and segmentation accuracy.
	\item[3)] An iterative teacher forcing scheme is devised to stabilize the training process and reduce the training difficulty.
	\item[4)] The proposed method has been evaluated on two open datasets and one in vivo in-house dataset and compared to six state-of-the-art methods. Results show that the proposed method possesses encouraging capabilities for simultaneous and accurate MR reconstruction and segmentation.
\end{enumerate}

The reminder of this paper is organized as follows: Section \ref{sec:method} describes our proposed method, Section \ref{sec:experiments and results} presents the experimental results and relevant analysis, and Section \ref{sec:discussion and conclusion} gives the discussion and conclusion.

\begin{figure*}[bp]
	\centering
	\includegraphics[width=1\linewidth]{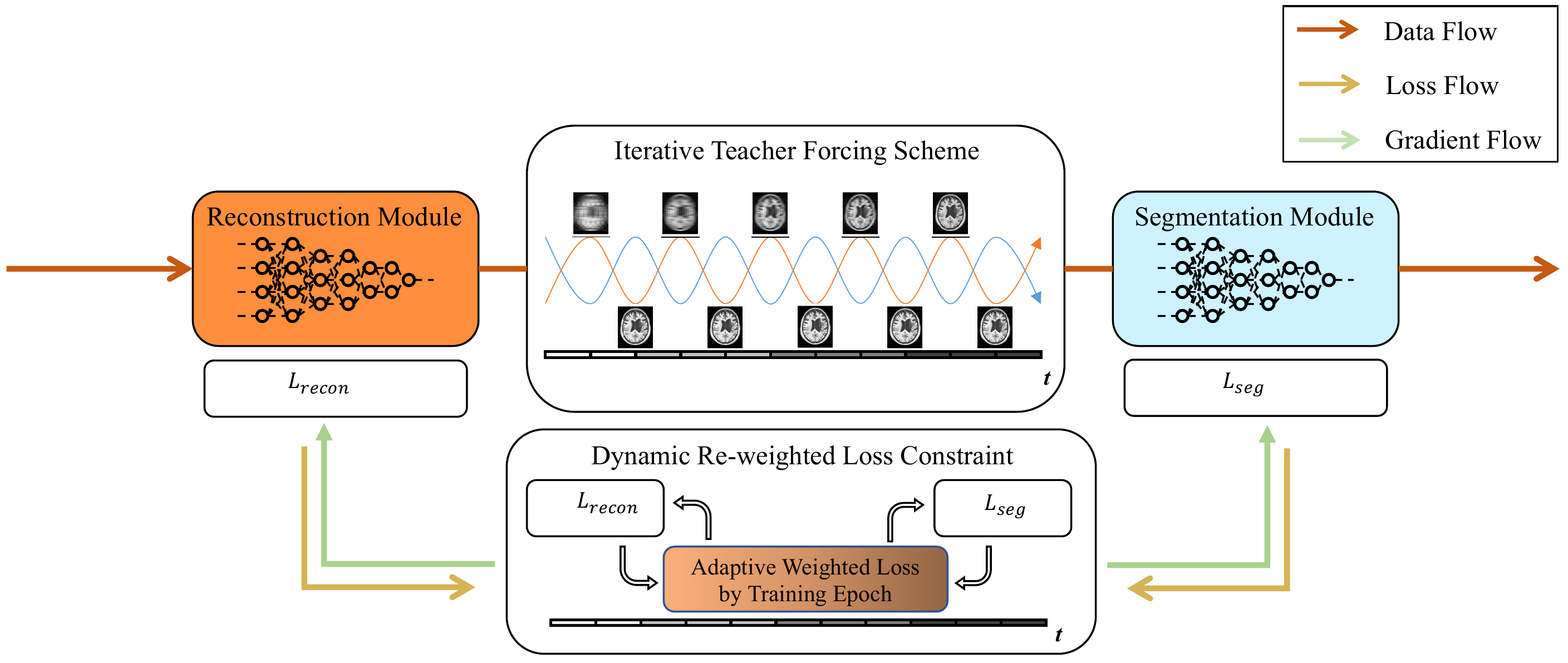}
	\caption{The overall workflow of the proposed multi-task MR imaging method. It consists of reconstruction and segmentation modules, which are D5C5 and U-Net respectively. Based on it, dynamic re-weighted loss constraint is designed to keep the contribution balance between reconstruction and segmentation tasks. Besides, an improved iterative teacher forcing scheme is devised to stabilize the training process and reduce the training difficulty of the dual-task network.}
	\label{Fig.1}
\end{figure*}

\section{Method}
\label{sec:method}
\subsection{Overview}
The overall workflow of the proposed multi-task MR imaging framework is shown in Fig. \ref{Fig.1}. It consists of reconstruction and segmentation modules. We selected D5C5 \cite{schlemper2017deep} as the reconstruction module and U-Net \cite{ronneberger2015u} as the segmentation module respectively. D5C5 \cite{schlemper2017deep} is a convolutional neural network (CNN) with a deep cascade structure. It integrated data consistency layers with convolutional layers in an end-to-end training manner for reconstruction. U-Net \cite{ronneberger2015u} with encoder-decoder structure based fully convolutional network (FCN) is a popular choice for medical image segmentation. The encoder and decoder were used to capture high-level semantic information and recover the spatial information respectively. In addition, multiple skip-connections were introduced between the encoder layer and the corresponding decoder layer to recover the information loss caused by down-sampling. The reason for selecting the published networks is that we focus on exploiting the multi-task imaging framework rather than proposing a reconstruction or segmentation network structure. The trained multi-task MR imaging method takes the under-sampled k-space data as input and outputs both the reconstructed images and the segmented masks simultaneously. 

\subsection{Dynamic Re-weighted Loss Constraint}
The proposed multi-task MR imaging framework was designed to produce high-quality reconstructed images and the segmented masks simultaneously. For the multi-task MR imaging method, the quality of the output of the downstream module was influenced by the upstream module greatly. The reason for it is that the input of the segmentation module is the output of the reconstruction module. Combining the segmentation module with the reconstruction module as a dual-task network could bridge the gap between two relational tasks. Multi-task imaging framework has the ability to make the reconstruction module keep the feature that is critical for segmentation instead of only the feature that is critical for visual performance. Commonly, this kind of network can be trained in an end-to-end manner with a loss function that is a fixed weighted summation of each task-specific loss function as shown in Eq. \eqref{Eq.1}:

\begin{align}
\label{Eq.1}
L=\sum_{i=1}^{N} \alpha_{i} L_{i}
\end{align}
where $L_{i}$ denotes the loss function of the $i$-th task and $\alpha_{i}$ denotes the corresponding weight. It makes the reconstruction module under the guidance of the loss function to keep the feature that can improve the quality of segmentation results. Such loss function has advantages including simple structure, easy implementation and low calculation amount. However, it can't guarantee the contribution balance between the two sub-tasks. Besides, this kind of loss function keeps the same weight for the training stage. In other words, the training process of the network will keep constant whether the particular sub-task is sufficiently optimized. It may be insufficient for producing high-quality reconstructed images and segmented masks simultaneously as the results shown in our experiments. Therefore, we hope that the weight of the task can be dynamically changed with the training process. It means that the center of the training process can be automatically transformed. Specifically, the center of the training process should be more inclined to optimize the reconstruction module instead of the segmentation module in the early training stage. In the later training stage, it is advisable to spend more efforts in optimizing the segmentation module when the loss of the reconstruction module is basically stable. It helps the network to keep the contribution balance of each task to ensure the quality of the final result. In summary, we proposed the following dynamic re-weighted loss constraint for the proposed network as shown in Eq. \eqref{Eq.2}:

\begin{align}
\label{Eq.2}
L=\alpha(t) L_{recon}+\beta(t) L_{seg}
\end{align}
where $L_{recon}$ and $L_{seg}$ denote the loss function of the reconstruction and segmentation tasks respectively. $\alpha(t)$ and $\beta(t)$ denote the corresponding weights, and $t$ denotes the corresponding training epoch. $\alpha(t)$ and $\beta(t)$ are the monotonic functions of $t$. In this paper, we selected $L_{2}$-norm as the loss function for reconstruction task and Cross-Entropy as the loss function for the segmentation task. The loss function of the reconstruction task is defined as Eq. \eqref{Eq.3}:

\begin{align}
\label{Eq.3}
L_{recon} = \| C(\mat{I}_{\mathrm{FS}})-\mat{I}_{\mathrm{US}}\|_{2}^{2}
\end{align}
where $\mat{I}_{\mathrm{FS}}$ and $\mat{I}_{\mathrm{US}}$ denote the images reconstructed from fully-sampled and under-sampled k-space data respectively, $C$ denotes the proposed network and $\|\cdot\|_{2}$ represents the $L_{2}$-norm. The loss function of the segmentation task is defined as Eq. \ref{Eq.4}:

\begin{align}
\label{Eq.4}
L_{seg} = 1.0 - \frac {2\times |\mat{y} .* \mat{\hat y}|}{|\mat{y}| + |\mat{\hat y}| + 0.1}
\end{align}
where $\mat{y}$ and $\mat{\hat y}$ denote the segmentation and annotation mask respectively, $.*$ denotes the pixel-wise multiplication of the two matrices and $|\cdot|$ indicates the element-wise sum operation.

\begin{figure}[!htbp]
	\centering
	\includegraphics[width=0.8\linewidth]{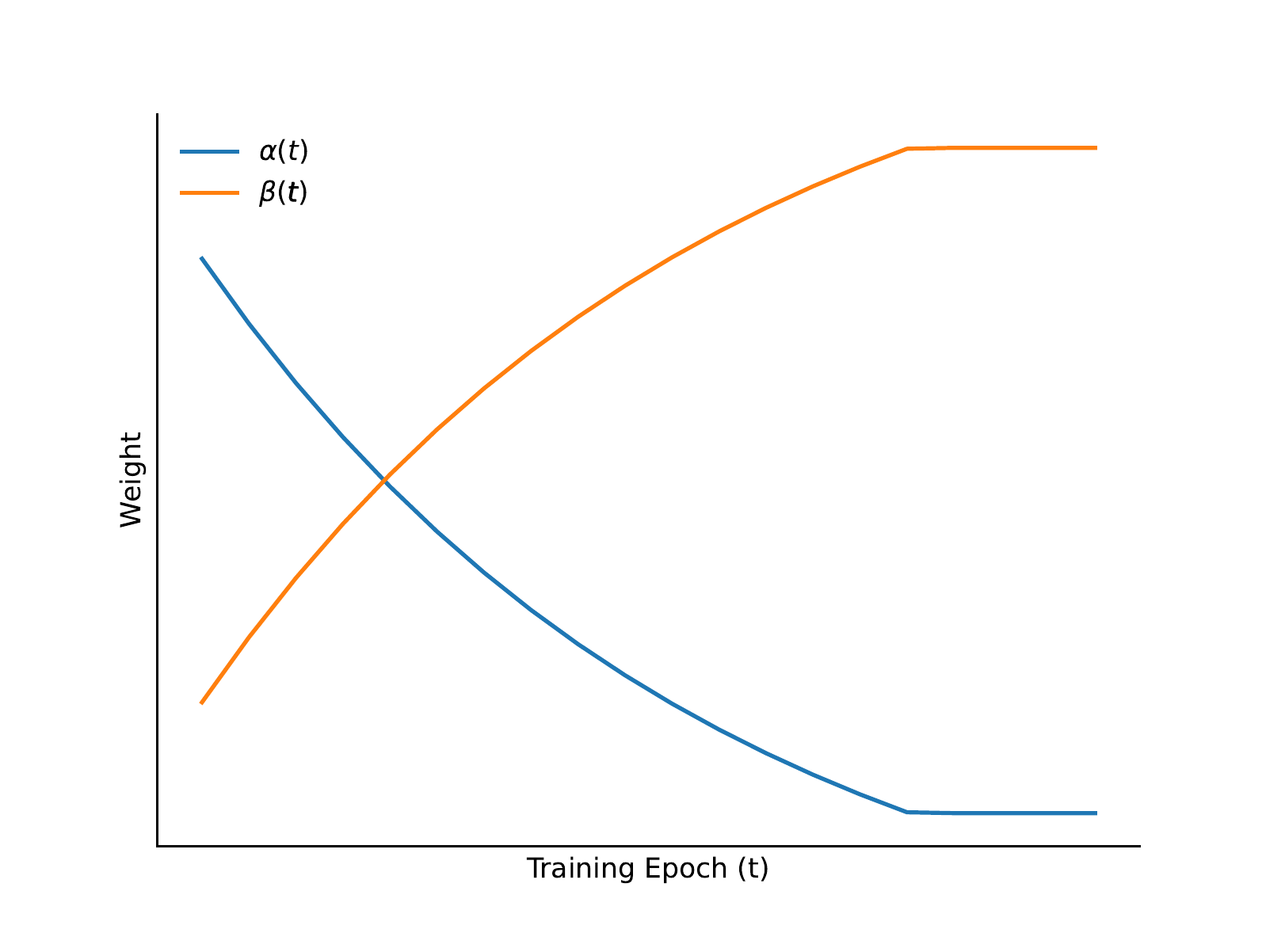}
	\caption{The change trend of $\alpha(t)$ and $\beta(t)$ with $t$.}
	\label{Fig.2}
\end{figure}

As shown in Fig. \ref{Fig.2}, $\alpha(t)$ increases while $\beta(t)$ decreases with the increase of $t$. In our proposed method, we set $\alpha(t)$ as the maximum value of the difference between the exponential transformation of $-t$ and $0.2$, and $0.05$ as shown in Eq. \eqref{Eq.5}. $\beta(t)$ was set as the difference between $1$ and $\alpha(t)$ as shown in Eq. \eqref{Eq.6}. It is designed to let the neural networks automatically adapt to the new weights according to their training stage. It was set to avoid over-large or over-small value of $\alpha(t)$ to make sure each task-specific loss is effective. The use of DRLC guarantees the proposed method can take into account both the reconstruction task and the segmentation task to produce high-quality reconstructed images and segmented masks simultaneously.

\begin{align}
\label{Eq.5}
\begin{array}{c}
\alpha(t)=\max \{\exp (-t)-0.2,0.05\}
\end{array}
\end{align}

\begin{align}
\label{Eq.6}
\begin{array}{c}
\beta(t)=1-\alpha(t)
\end{array}
\end{align}

\subsection{Iterative Teacher Forcing Scheme}
In the training phase, supposing there is a serialization task with $T$ steps. Let $x_{t}$, $y_{t}$  and $\hat{y}_{t}$ denote the input, ground truth and output of the step $t$ respectively. For common auto-regressive mode, $x_{t}$ is given with $\hat{y}_{t-1}$. The limitation of it is that the error of low-level models will be accumulated gradually even to make the network divergent. For teacher forcing mode, $y_{t-1}$ is taken as the input of the step $t$ instead of $\hat{y}_{t-1}$. It is equivalent to take the ground truth as the teacher to correct the output of low-level models in time. The benefit provided by the use of the teacher forcing scheme is making the training process stable and quick to reduce the training difficulty of the network. However, the teacher forcing scheme may lead to bias exposure. It refers to the performance gap between the training and testing processes. Namely, the use of the teacher forcing scheme makes the inputs of the step $t$ of the network in the training process and the testing process are inferred from different distributions. It may lead to discrepancy between the decoding actions of the network in the training and testing phases. Meanwhile, the constraint provided by the teacher forcing scheme may over-correct the output to reduce the representation ability of the network.

\begin{figure*}[!htbp]
	\centering
	\includegraphics[width=1\linewidth]{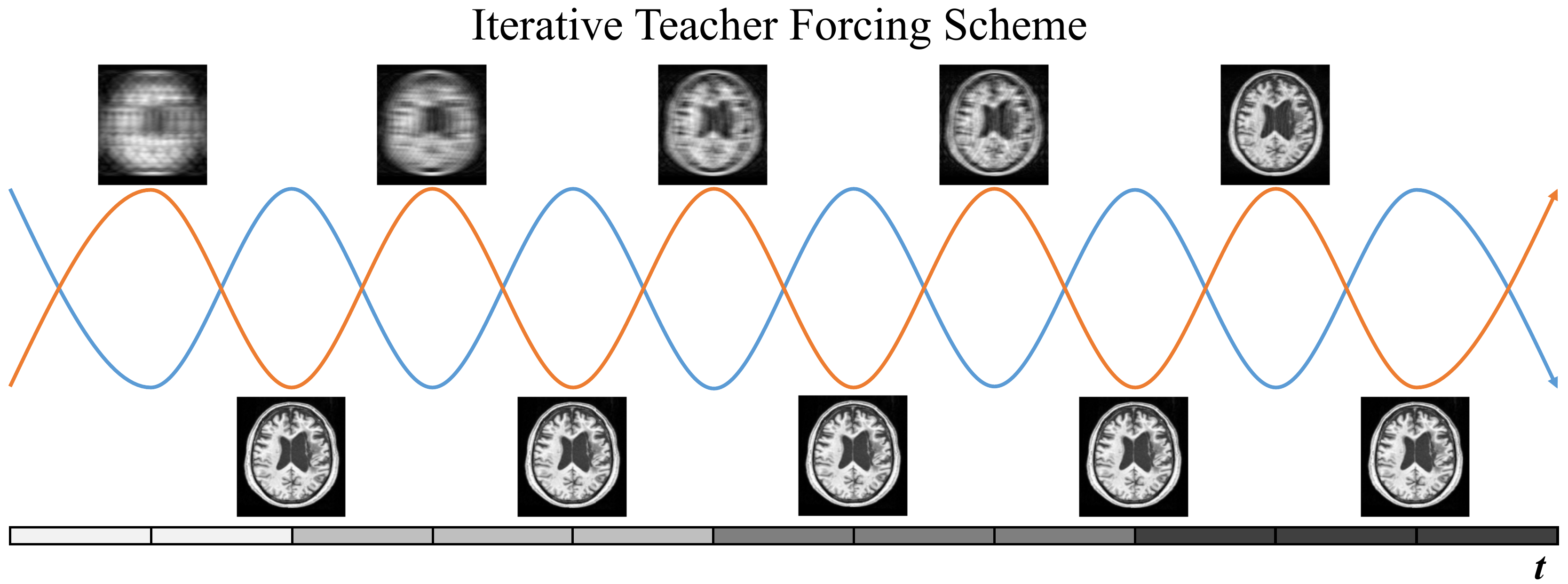}
	\caption{The difference between the original teacher forcing scheme and the proposed iterative teacher forcing scheme is that we take the ground truth denoted by the blue line and the actual output denoted by the orange line of step $t-1$ as the input of step $t$ alternately instead of the fixed ground truth of step $t-1$. It is beneficial for improving the training process while avoiding exposure bias and over-correction issue.}
	\label{Fig.3}
\end{figure*}

To overcome the aforementioned limitations of the original teacher forcing scheme, we propose an iterative teacher forcing scheme. As shown in Fig. \ref{Fig.3}, we use the blue line to denote the actual output of step $t-1$ and the orange line to denote the ground truth of step $t-1$ respectively. The difference between the original teacher forcing scheme and the proposed iterative teacher forcing scheme is that taking the ground truth and the actual output of step $t-1$ as the input of step $t$ alternately rather than the fixed ground truth of step $t-1$. The proposed iterative teacher forcing scheme not only can correct the prediction of the low-level model to a certain extent but also can avoid the performance gap between the training and testing processes. 

Specifically, we treated multi-task MR imaging as a two-step serialization task that begins from the reconstruction step to the segmentation step for applying the iterative teacher forcing scheme. As far as we know, it is the first time that the teacher forcing scheme has been used in neural networks outside RNN. In the training phase of the proposed network, we took the under-sampled k-space data as the input of the first step $x_{1}$. The output $\hat{y}_{1}$ and the ground truth $y_{1}$ of the first step were selected as the input of the second step $x_{2}$ iteratively. It prevents the error accumulation of the reconstruction step from propagating to the segmentation step. In addition, the use of teacher forcing scheme is beneficial for stabilizing the training process, reducing the training difficulty and protecting the presentation ability of the proposed network.

\section{Experiments and Results}
\label{sec:experiments and results}
\subsection{Experimental Setting}
\subsubsection{Data}
We instantiated our method on two competitive MR image datasets including ATLAS \cite{liew2018atlas} and realistic simulated k-space data generated from Brainweb \cite{cocosco1997brainweb}. The ATLAS dataset consists of 229 intensity normalized subjects on T1 modality in standard space (normalized to the MNI-152 template) collected from 11 cohorts worldwide, with an in-plane resolution of $1mm^3$, in which 120 subjects were used for training, 40 for validation and 69 for testing. 20 cases were employed from the publicly available simulated brain dataset, Brainweb, where 12 brain tissue types were included with known distributions, in which 17 cases were used for training and 3 for testing.  

We also conducted our approach on an in house in-vivo dataset. It consists of 40 cases and 485 slices from FLAIR sequence obtained with a Siemens 1.5T scanner. The FOV phase amplitude is $70\% - 87.5\%$. The slice thickness ranges from 6.0 mm to 7.0 mm and the slice interval ranges from 7.8 mm to 9.1 mm. Lesions  were annotated by experienced clinicians using ITK snap. 20 cases (247 slices) are used for training with the rest 20 cases (238 slices) for testing. 

\begin{figure}[!htbp]
	\centering
	\includegraphics[width=0.6\linewidth]{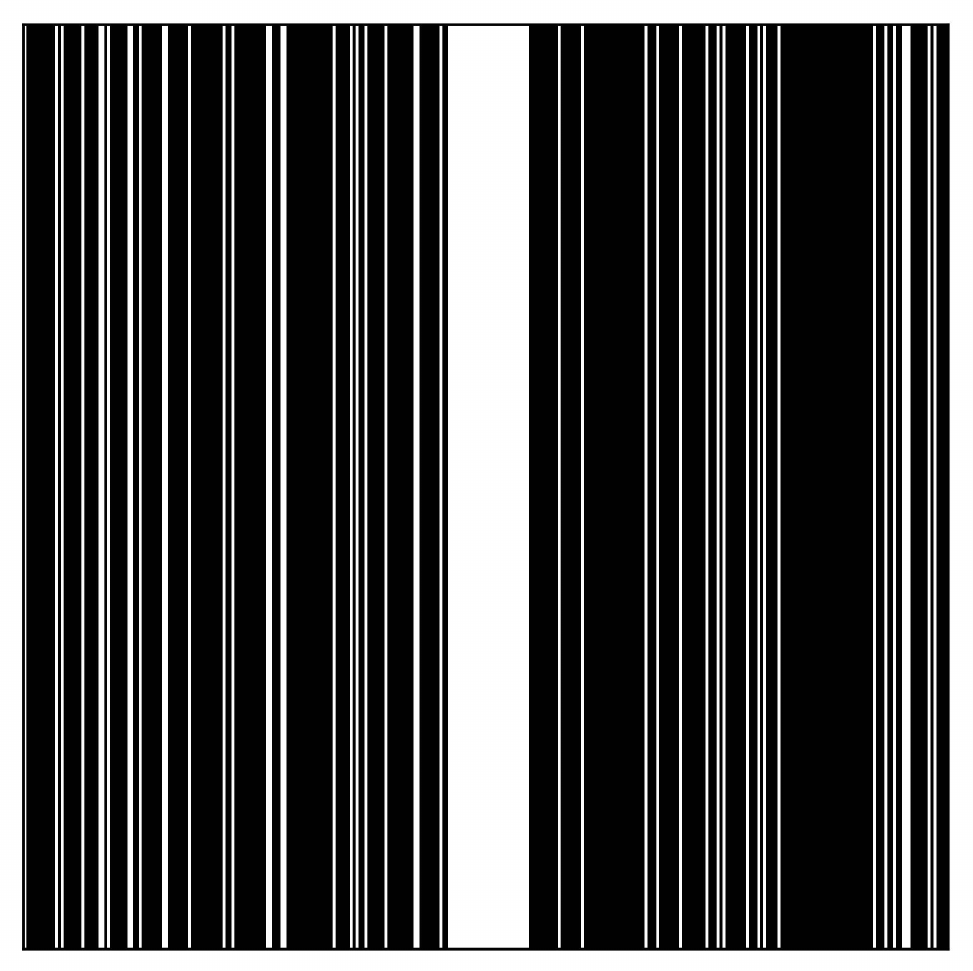}
	\caption{Examples of under-sampled k-space trajectories.}
	\label{Fig.4}
\end{figure}

For k-space data sub-sampling, we performed masks as shown in Fig. \ref{Fig.4} on the fully-sampled acquisition retrospectively. For a volume, all slices were applied to the same under-sampling mask. The overall acceleration factor was set as four. To achieve the desired acceleration factor, $8\%$ of all k-space lines from the central region were kept and the remaining k-space lines were kept with a set probability uniformly at random. It was set to meet the general conditions for compressed sensing \cite{candes2006robust,lustig2007sparse}.

\begin{table*}[bp]
	\caption{Ablation analysis of iterative teacher forcing scheme and dynamic re-weighted loss constraint on ATLAS dataset. For each metric, we mark the best in bold.}
	\label{Table.1}
	\centering
	\begin{tabular}{ccccccc}
		\toprule
		ITFS & DRLC &      Dice      &    Precision   &      Recall     &      PSNR      &      SSIM      \\ \midrule
		&                  &      $0.494\pm 0.071$      &      $0.636\pm 0.135$     &      $0.462\pm 0.067$      &      $28.78\pm 0.732$     &      $0.957\pm 0.005$     \\ 
		\checkmark      &                  &      $0.499\pm 0.096$     &      $0.682\pm 0.109$     &      $0.467\pm 0.024$      &      $27.20\pm 0.533$     &      $0.943\pm 0.004$     \\ 
		&    \checkmark    &     $0.510\pm 0.068$           &     $0.686\pm 0.126$           &      $0.472\pm 0.060$           &      $28.84\pm 0.619$          &         $0.957\pm 0.005$       \\ 
		\checkmark      &    \checkmark    & $\textbf{0.518}\pm 0.064$          & $\textbf{0.688}\pm 0.076$ & $\textbf{0.479}\pm 0.020$ & $\textbf{28.87}\pm 0.409$ & $\textbf{0.958}\pm 0.00005$     \\ \bottomrule
	\end{tabular}
\end{table*}

\begin{table*}[bp]
	\caption{Ablation analysis of different dynamic re-weighted loss constraint. For each metric, we mark the best in bold.}
	\label{Table.2}
	\centering
	\begin{tabular}{cccccc}
		\toprule
		Re-weighted Loss Strategy                                            & Dice           & Precision      & Recall         & PSNR           & SSIM           \\ \midrule
		\begin{tabular}[c]{@{}c@{}}Fixed Weight\\ ($\alpha=0.5$ and $\beta=0.5$)\end{tabular} & $0.494\pm 0.083$          & $0.673\pm 0.104$          & $0.459\pm 0.045$         & $27.23\pm 0.620$          & $0.933\pm 0.009$          \\ 
		\begin{tabular}[c]{@{}c@{}}Fixed Weight\\ ($\alpha=0.2$ and $\beta=0.8$)\end{tabular} & $0.485\pm 0.077$          & $0.662\pm 0.080$          & $0.444\pm 0.042$          & $28.70\pm 0.519$          & $0.955\pm 0.007$          \\ 
		\begin{tabular}[c]{@{}c@{}}Dynamic Weight\\ (Linear)\end{tabular}    &  $ 0.494\pm 0.073$          & $0.664\pm 0.087$          & $0.452\pm 0.031$          & $28.87\pm 0.516$          & $0.956\pm 0.0006$          \\ 
		\begin{tabular}[c]{@{}c@{}}Dynamic Weight\\ (Exponent)\end{tabular}  & $\textbf{0.518}\pm 0.064$          & $\textbf{0.688}\pm 0.076$ & $\textbf{0.479}\pm 0.020$ & $\textbf{28.87}\pm 0.409$ & $\textbf{0.958}\pm 0.00005$     \\ \bottomrule
	\end{tabular}
\end{table*}

\subsubsection{Implementation Details}
This subsection presented some of the details regarding the sub-sampling procedure. The model was implemented using Pytorch deep learning framework. D5C5 \cite{schlemper2017deep} was selected as the reconstruction module and U-Net \cite{ronneberger2015u} as the segmentation module. The model was trained for 50 epochs using Adam \cite{kingma2014adam} optimization algorithm on a single NVIDIA Titan Xp GPU.

For ATLAS dataset, 189 slices were collected for every subject. Each slice was cropped to $224\times 192$ for the convenience of deep learning model inference. The batch size was set to 16, and the initial learning rate for both sub-modules was set to $1e-6$ with a decreasing rate of 0.2 for every 10 epochs. 229 subjects were randomly split into 3 sub-datasets for training, validation and testing, respectively. Each of these three sub-datasets contained 120, 40 and 69 subjects, respectively. Segmented brain lesions were employed to calculate the segmentation metrics. The model was trained for 50 epochs using Adam optimizer.

For simulated Brainweb data, 20 cases were employed with known tissue annotation, 17 for training and 3 for testing. 12 tissue types were given, in which 3 of them (CSF, WM, and GM) were used to calculate the segmentation quantitative metrics. All the 3D images were resized and cropped to the size of $362\times 416\times 352$. The batch size was set to 4, and the initial learning rate was set to $1e-4$. Random flip, rotation and affine transformation are used as data augmentation methods. The model was trained for 3000 epochs using Adam optimizer.

For in vivo in-house dataset, the MR data were collected by Guizhou Provincial People's Hospital. Data from 40 cases were obtained with FLAIR sequence and the stroke lesions were annotated by experienced clinicians using ITK snap, with 20 cases as the training set and the rest 20 cases as the test set. All images were obtained with a Siemens 1.5T scanner. The FOV phase amplitude is $70 \% -87.5 \% $. The slice thickness ranges from 6.0 mm to 7.0 mm and the slice interval ranges from 7.8 mm to 9.1 mm. All the data were resized to $224\times 224$ with intensity normalized. The batch size was set to 16, and the initial learning rate was set to $1e-4$. The model was trained for 50 epochs using Adam optimizer.

\subsection{Quantitative Evaluation Metrics}
For the ATLAS dataset and in vivo in-house dataset, five metrics are used to measure the performance of the segmentation and reconstruction results, including the Dice score, precision, recall, peak signal-to-noise ratio (PSNR) and structural similarity index matrix (SSIM). The former three metrics were employed to evaluate the segmentation performance and the latter two were used to evaluate the reconstruction performance. For the Brainweb dataset, five metrics are used to measure the performance of the segmentation and reconstruction results, including the Dice score on CSF, GM, WM, PSNR and SSIM. All the metrics referred to the 3D predicting results. Average metric scores on the testing dataset were listed for the comparison between different methods.

\subsection{Ablation Study}
\subsubsection{Effectiveness of ITFS and DRLC}
\begin{figure}[!htbp]
	\centering
	\includegraphics[width=1\linewidth]{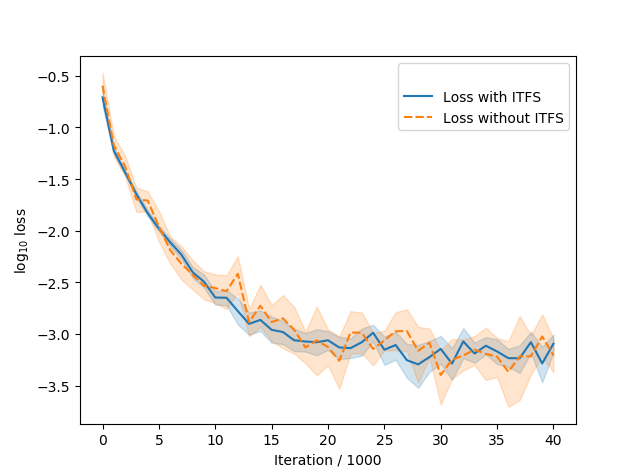}
	\caption{The loss values with or without ITFS. The colored area covers $95\%$ of the loss values. It is clear that using ITFS could make the training process more stable.}
	\label{Fig.ITFS}
\end{figure}

In this section, we focus on evaluating the effectiveness of the proposed modules of ITFS and DRLC. First, we trained a joint framework of reconstruction and segmentation modules from scratch without the iterative teacher forcing scheme and dynamic re-weighted loss constraint. Second, we alternatively introduced the iterative teacher forcing scheme and dynamic re-weighted loss constraint to the joint framework. The related quantitative comparison results are shown in Table \ref{Table.1}. We could see both the iterative teacher forcing scheme and dynamic re-weighted loss constraint improved the performance of the proposed network. On the basis of it, we leveraged the iterative teacher forcing scheme and dynamic re-weighted loss constraint simultaneously. It is clearly observed that the network equipped with both the iterative teacher-forcing scheme and the dynamic re-weighted loss constraint achieved better performance than other networks. They were beneficial for producing high-quality reconstruction and segmentation results simultaneously. Furthermore, the results shown in Table \ref{Table.1} demonstrated that the dynamic re-weighted loss constraint contributed more to the performance gain than the iterative teacher-forcing does. However, ITFS has a more important capability of stabilizing the training process. As shown in Fig. \ref{Fig.ITFS}, the introduction of ITFS improved the convergence of the proposed network. It stabilized the training process of the dual-task network to reduce the training difficulty.

\subsubsection{Ablation Analysis of Different Kinds of DRLC}
As the results showed above, the dynamic re-weighted loss constraint helped the network produce promising results. In this section, we compared the performance of different re-weighted loss constraints to validate the effectiveness of our proposed exponential-like dynamic re-weighed loss constraint. As shown in Table \ref{Table.2}, we have evaluated three kinds of weight setting schemes, namely the fixed weight, dynamic weight with linear settings and our proposed scheme. Compared to the fixed weights of $\alpha$ and $\beta$, dynamic re-weighted loss constraint achieved better performance for both reconstruction and segmentation.  Dynamic self-adopted loss adjustment helps the loss value adapt to the training stage, which avoids too large loss and gradient value at the final stage of training. Furthermore, the performance of the re-weighted loss based on exponential operations was better than that of the re-weighted loss based on linear functions. This is intuitively explainable since the first few epochs may change drastically while it becomes steady with respect to the iterations. Therefore, we selected the re-weighted loss with exponential formations in the following experiments. 

\subsection{Evaluation and comparison on ATLAS dataset}
\begin{table*}[tp]
	\caption{Comparison experiments results of different methods on ATLAS dataset. For each metric, we mark the best in bold.}
	\label{Table.3}
	\centering
	\begin{tabular}{cccccc}
		\toprule
		Method                          &      Dice      &    Precision   &     Recall     &      PSNR      &      SSIM          \\ \midrule
		SynNet                          &     $0.357\pm 0.072$      &     $0.459\pm 0.085$      &     $0.352\pm 0.068$      &       -        &       -            \\ 
		LI-Net                          &     $0.312\pm 0.067$      &     $0.478\pm 0.095$      &     $0.278\pm 0.055$      &       -        &       -            \\ 
		SERANet                         &     $0.213\pm 0.064$      &     $0.656\pm 0.179$      &     $0.149\pm 0.034$      &       -        &       -            \\ 
		SegNetMRI                       &     $0.224\pm 0.067$      &     $0.618\pm 0.164$      &     $0.156\pm 0.034$      &     $28.22\pm 0.746$      &     $0.950\pm 0.006$          \\ 
		\begin{tabular}[c]{@{}c@{}}U-Net\\ (full-sample)\end{tabular} &     $0.512\pm 0.085$      &     $0.632\pm 0.102$      &     $\textbf{0.511}\pm 0.094$      &       -        &       -            \\ 
		D5C5                                   &      -         &       -        &       -        &     $28.78\pm 0.689$      &     $0.958\pm 0.0005$         \\ 
		ours                            & $\textbf{0.518}\pm 0.064$          & $\textbf{0.688}\pm 0.076$ & $0.479\pm 0.020$ & $\textbf{28.87}\pm 0.409$ & $\textbf{0.959}\pm 0.00005$     \\ \bottomrule
	\end{tabular}
\end{table*}

\begin{figure}[bp]
	\centering
	\includegraphics[width=1\linewidth]{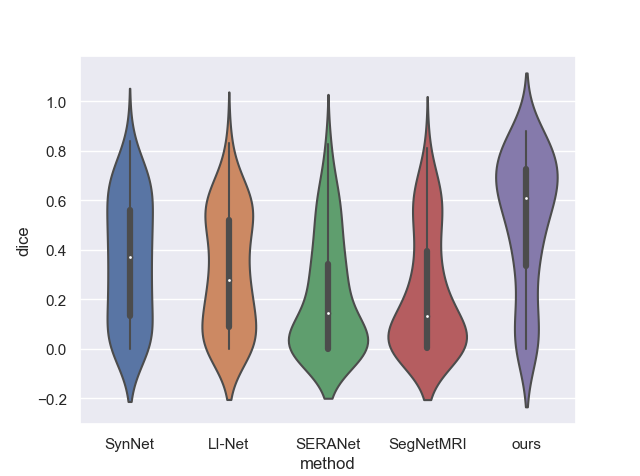}
	\caption{Violin-plot of comparison of segmentation results on ATLAS dataset.}
	\label{Fig.violinplot}
\end{figure}

\begin{figure}[!htbp]
	\centering
	\includegraphics[width=1\linewidth]{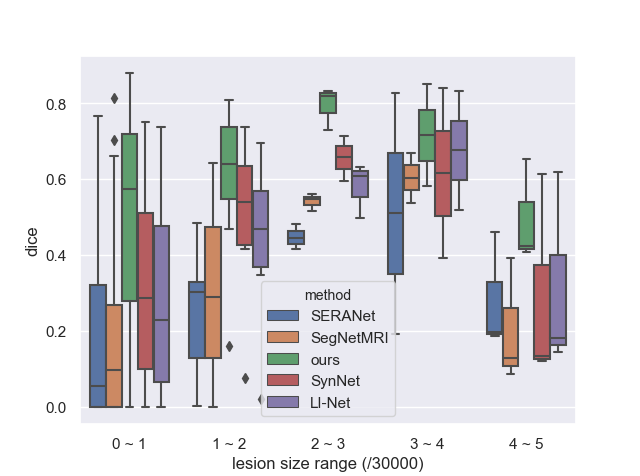}
	\caption{Boxplot of comparison of segmentation results on ATLAS dataset.}
	\label{Fig.boxplot}
\end{figure}

To validate the effectiveness of our method, we conducted experiments on ATLAS dataset, which is publicly available and offers a high variety of lesions. Noting that, our proposed method and SegNetMRI could provide reconstructed images and segmented masks simultaneously. However, SynNet, LI-Net, and SERANet just could provide segmented masks. Table \ref{Table.3} and Fig. \ref{Fig.violinplot} show the comparison of reconstruction and lesion segmentation performance on ATLAS dataset. We compared our method with existing segmentation and reconstruction from under-sampled k-space data approaches. We can see that our approach achieved better performance on lesion segmentation and promising performance on image reconstruction, with more that 0.15 gain on Dice score. It is worth noticing from Fig. \ref{Fig.violinplot} that our approach achieved smaller variety on Dice score, which indicates a better segmentation performance on small lesions. Segmentation results are encouraging even compared with segmentation from fully-sampled data. 

To have a comprehensive comparison with existing segmentation from under-sampled data methods, we reported the box-plot on Dice score for different lesion sizes in Fig. \ref{Fig.boxplot}. This figure shows that all of the methods perform well on median size lesions (lesion size 2 and 3). However, our approach achieves encouraging results on small (lesion size 0 and 1) and huge (lesion size 5) lesions, while other methods perform worse. The promising segmentation result suggests the effectiveness and good generalization capability of the proposed method. 

\begin{figure*}[bp]
	\centering
	\includegraphics[width=1\linewidth]{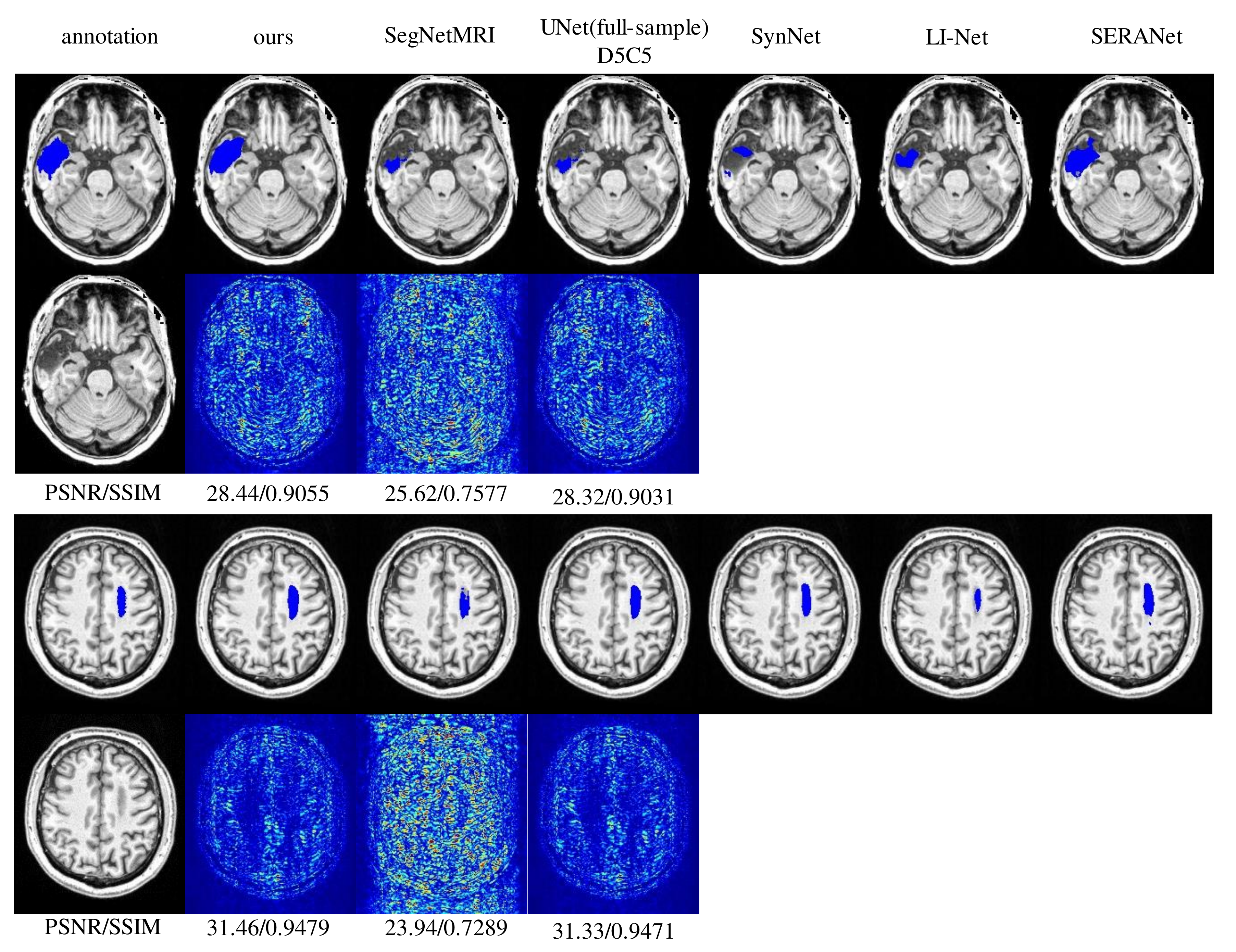}
	\caption{Examples of our lesion segmentation results and image reconstruction error maps on ATLAS dataset. From left to right: label, ours, SegNetMRI, U-Net (full-sample) (top) or D5C5 (bottom), SynNet, LI-Net, SERANet.}
	\label{Fig.SegReconATLAS}
\end{figure*}

The quantitative comparison of the reconstruction performance of different networks is shown in Table \ref{Table.3}. As we can see, our proposed method obtained promising quantitative metrics compared with other methods. Besides, we selected two slices to show the reconstruction performance of different networks, which are shown in Fig. \ref{Fig.SegReconATLAS}. As we can see in Fig. \ref{Fig.SegReconATLAS}, all the methods were able to generate high-quality images. The suppression of artifacts improved the visual quality of the reconstruction images greatly. Nevertheless, there were differences between the reconstruction results of different networks. Reference to the error maps and quantitative metrics, the reconstruction ability of the proposed method was better than that of the other method.

\begin{figure*}[tp]
	\centering
	\includegraphics[width=1\linewidth]{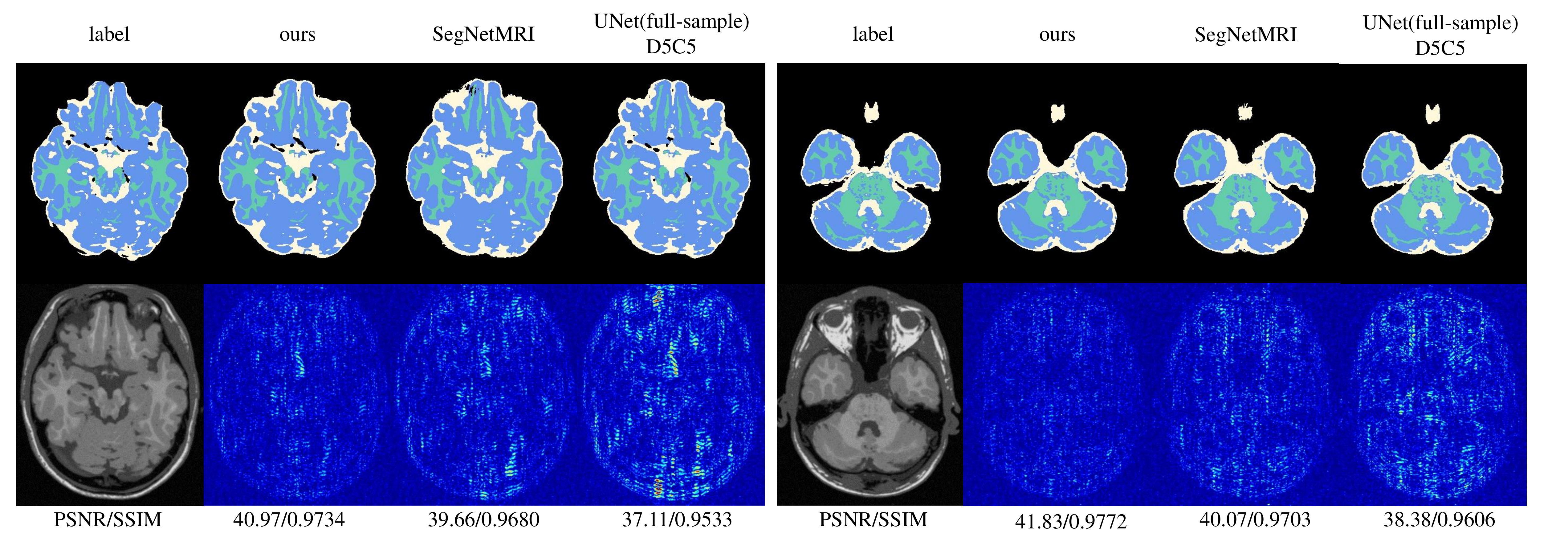}
	\caption{Examples of our lesion segmentation results and image reconstruction error maps on simulated Brainweb data. From left to right: label, ours, SegNetMRI, U-Net (full-sample) (top) or D5C5 (bottom).}
	\label{Fig.Brainweb}
\end{figure*}

\begin{figure*}[tp]
	\centering
	\includegraphics[width=1\linewidth]{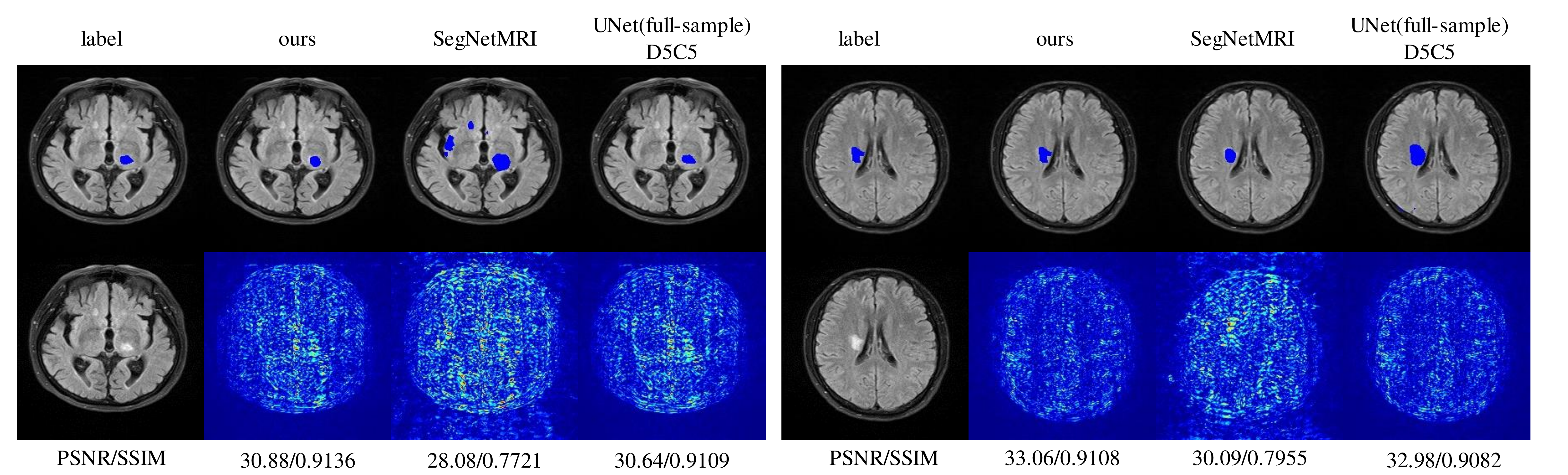}
	\caption{Examples of our lesion segmentation results and image reconstruction error maps on in vivo data. From left to right: label, ours, SegNetMRI, U-Net (full-sample) (top) or D5C5 (bottom).}
	\label{Fig.7}
\end{figure*}

\subsection{Evaluation and comparison on Simulated Brainweb Data}
To validate the effectiveness and robustness of the proposed method, we conduct experiments on simulated Brainweb data, which is publicly available and offers digital brain phantoms with 12 tissue types. Three tissues were used to evaluate the segmentation performance because of their large size. Table \ref{Table.brainweb} and Fig. \ref{Fig.Brainweb} shows the comparison of the image reconstruction and tissue segmentation results. We can see that our method achieved better performance on CSF segmentation and PSNR, with 0.04 and 1.2 dB improvement on CSF Dice score and PSNR, respectively. Our method also achieved encouraging performance on WM and GM segmentation and SSIM, with 0.9285, 0.9354 and 0.993, respectively.

\subsection{Evaluation and comparison on in vivo in-house Dataset}
To validate the effevtiveness of the proposed on in vivo data, we conduct experiments on in vivo in-house dataset. Table \ref{Table.5} and Fig \ref{Fig.7} shows the comparison of the image reconstruction and lesion segmentation results. We can conclude that our approach achieved encouraging performance for both reconstruction and segmentation, with 0.864 in Dice, 32.65 dB in PSNR and 0.983 in SSIM respectively. As we can see in Figure \ref{Fig.7}, all the methods achieves promising reconstruction error maps and segmentation masks. Compared with the error maps and segmentation masks achieved by other methods, our approach was more effective in these two tasks.

\begin{table*}[!htbp]
	\caption{Comparison experiments results of different methods on simulated Brainweb dataset. For each metric, we mark the best in bold.}
	\label{Table.brainweb}
	\centering
	\begin{tabular}{cccccc}
		\toprule
		Method                & CSF            & WM             & GM             & PSNR            & SSIM           \\ \midrule
		SegNetMRI             & $0.792\pm 0.044$          & $0.915\pm 0.008$          & $0.925\pm 0.021$          & $41.48\pm 0.077$           & $0.991\pm 0.001$          \\ 
		D5C5                  & -              & -              & -              & $39.68\pm 0.057$            & $0.993\pm 0.001$          \\ 
		\begin{tabular}[c]{@{}c@{}}U-Net\\ (full-sample)\end{tabular}      
		& $0.862\pm 0.013$           & $0.928\pm 0.002$ & $0.933\pm 0.001$          & -                & -          \\ 
		ours                 & $\textbf{0.869}\pm 0.003$  & $\textbf{0.958}\pm 0.001$          & $\textbf{0.960}\pm 0.001$          & $\textbf{42.68}\pm 0.044$   & $\textbf{0.997}\pm 0.001$ \\ \bottomrule
	\end{tabular}
\end{table*}

\begin{table*}[!htbp]
	\caption{Comparison experiments results of different methods on in-house dataset. For each metric, we mark the best in bold.}
	\label{Table.5}
	\centering
	\begin{tabular}{cccccc}
		\toprule
		Method        & Dice            & precision       & recall          & PSNR           & SSIM           \\ \midrule
		ours          & $\textbf{0.864}\pm 0.001$  & $\textbf{0.880}\pm 0.011$          & $\textbf{0.868}\pm 0.031$          & $\textbf{32.65}\pm 0.013$ & $\textbf{0.986}\pm 0.001$          \\ 
		SegNetMRI     & $0.841\pm 0.004$          & $0.814\pm 0.024$          & $0.859\pm 0.033$ & $32.52\pm 0.057$          & $0.983\pm 0.006$          \\
		\begin{tabular}[c]{@{}c@{}}U-Net\\ (Full-sample)\end{tabular} 
		& $0.854\pm 0.015$          & $0.830\pm 0.007$ & $0.7887\pm 0.107$          & -              & -              \\
		D5C5                                                          & -               & -               & -               & $32.49\pm 0.044$ & $0.984\pm 0.008$ \\ \bottomrule
	\end{tabular}
\end{table*}

\section{Discussion and Conclusion}
\label{sec:discussion and conclusion}
In this study, we propose a multi-task MR imaging framework with iterative teacher forcing scheme and dynamic re-weighted loss constraint. The promising reconstruction and segmentation results were achieved through our proposed re-weighted deep learning framework. As the comparison experiments showed, the proposed iterative teacher forcing scheme and dynamic re-weighted loss constraint co-prompt the quality of both the reconstruction task and segmentation task. The main drawback of directly combing the loss function of the corresponding task is that may lead to an unbalance between the performance of two modules. The proposed dynamic re-weighted loss constraint has the ability to control the training process of the network to take two considerations on both reconstruction and segmentation tasks. It helps the network avoid the over-large or over-small value of gradient in the training process to make sure each task-specific loss effective. In other words, the core of the training process of the proposed network was managed dynamically. Furthermore, the error of the reconstruction task may influence the quality of the segmentation task. Although we can leverage the teacher forcing scheme to solve the problem mentioned above, the incident exposure bias and over-correction may breakdown the performance of the network. The proposed iterative teacher forcing scheme not only internally correct the prediction of the low-level model but also help avoiding the occurrence of exposure bias and over-corrections for the reconstruction model. It makes sure the quality of the segmentation module not influenced by the reconstruction error as much as possible. Therefore, we argue that the segmentation performance gain mainly came from the iterative teacher forcing scheme. Moreover, compared with existing under-sampled k-space MR image segmentation methods, the proposed method segment small lesions more efficiently, which is crucial in clinical practice. We demonstrate that this performance gains mainly due to the effectiveness of the task-driven reconstruction model, which focuses on lesion region reconstruction. In the future, we consider to explore our method for dynamic MR imaging.  Another key that should be explored is the potential training cost of the iterative teacher forcing scheme and dynamic re-weighted loss constraint. Alternatives will be explored in the future for slighter training costs. 

\section{Acknowledgments}
\label{sec:acknowledgments}
This research was partly supported by Scientific and Technical Innovation 2030 - "New Generation Artificial Intelligence"  Project (2020AAA0104100, 2020AAA0104105), the National Natural Science Foundation of China (61871371, 81830056), Key-Area Research and Development Program of GuangDong Province (2018B010109009), the Basic Research Program of Shenzhen (JCYJ20180507182400762), Youth Innovation Promotion Association Program of Chinese Academy of Sciences (2019351).
\bibliographystyle{IEEEtran}
\bibliography{mylib}
\end{document}